\documentclass[10pt,twocolumn,letterpaper]{article}

\usepackage{iccv}
\usepackage{times}
\usepackage{epsfig}
\usepackage{graphicx}

\usepackage[pagebackref=true,breaklinks=true,letterpaper=true,colorlinks,bookmarks=false]{hyperref}

\usepackage[numbers]{natbib}
\usepackage{multicol}

\usepackage{hyperref}       
\usepackage{url}            
\usepackage{booktabs}       
\usepackage{amsfonts}       
\usepackage{nicefrac}       
\usepackage{microtype}      
\usepackage{graphicx}
\usepackage{amsmath} 
\usepackage{amssymb}  
\usepackage{gensymb}
\usepackage{url}
\usepackage{xcolor}
\usepackage{multirow}
\usepackage{hyperref}
\usepackage{paralist}

\usepackage{color}
\definecolor{darkred}{rgb}{0.5, 0.0, 0.0}
\definecolor{darkblue}{rgb}{0.0, 0.0, 1.0}
\newcommand{\hide}[1]{}

\newcommand{\cutabstractup}{\vspace*{-0.2in}}

\newcommand{\cutsectionup}{\vspace*{-0.05in}}
\newcommand{\cutsectiondown}{\vspace*{-0.05in}}

\newcommand{\cutparagraphup}{\vspace*{-0.1in}}

\usepackage{tikz}
\definecolor{g-red}{HTML}{DB4437}
\definecolor{g-blue}{HTML}{4285F4}
\definecolor{g-green}{HTML}{0F9D58}
\definecolor{g-yellow}{HTML}{F4B400}
\definecolor{g-orange}{HTML}{FF9800}
\definecolor{g-grey}{HTML}{9E9E9E}

\usepackage[colorinlistoftodos,textsize=footnote]{todonotes}

\newcounter{hl}

\def\iccvPaperID{2007} 

\iccvfinalcopy
\ificcvfinal\fi

\begin{document}

\title{Data-Efficient Learning for Sim-to-Real Robotic Grasping \\ using Deep Point Cloud Prediction Networks}

\author{Xinchen Yan
\thanks{Work done during internship with Google Brain and X.}
\\University of Michigan
\and
Mohi Khansari
\\X Inc.
\and
Jasmine Hsu
\\Google Brain
\and
Yuanzheng Gong
\\X Inc.
\and
Yunfei Bai
\\X Inc.
\and
S\"oren Pirk
\\Google Brain
\and
Honglak Lee
\\Google Brain
}



\maketitle
\begin{abstract}
Training a deep network policy for robot manipulation is notoriously costly and time consuming as it depends on collecting a significant amount of real world data. To work well in the real world, the policy needs to see many instances of the task, including various object arrangements in the scene as well as variations in object geometry, texture, material, and environmental illumination. 
In this paper, we propose a method that learns to perform table-top instance grasping of a wide variety of objects while using \textbf{no real world grasping data}, outperforming the baseline using 2.5D shape by 10\%. Our method learns 3D point cloud of object, and use that to train a domain-invariant grasping policy. We formulate the learning process as a two-step procedure: 1) Learning a domain-invariant 3D shape representation of objects from $\sim$76K episodes in simulation and $\sim$530 episodes in the real world, where each episode lasts less than a minute and 2)~Learning a critic grasping policy in simulation only based on the 3D shape representation from step~1. Our real world data collection in step~1 is both cheaper and faster compared to existing approaches as it only requires taking multiple snapshots of the scene using a RGBD camera. 
Finally, the learned 3D representation is not specific to grasping, and can potentially be used in other interaction tasks. 
Video demos can be accessed via {\color{blue}\url{https://sites.google.com/site/shapeawaresimtoreal/}}.

\end{abstract}
\cutabstractup

\cutsectionup
\section{Introduction}
\cutsectiondown

Learning a domain-invariant representation for object manipulation in real-world environments with minimum supervision is a fundamental challenge in vision and robotics.
State-of-the-art learning systems rely on collecting large-scale datasets with interaction labels (e.g., success or failure) using multiple robots (e.g., KUKA robotic arms) to update the model parameters in parallel via end-to-end training~\cite{levine2016learning}. 
The cost of deploying these learning systems to a new setting (e.g. a novel task under different environment) is very high and the efficiency is limited by the number of robots that can be deployed in parallel. 
Therefore, existing learning based methods also use simulation data to alleviate the data collection problem. However, the existence of the sim-to-real gap imposes additional difficulty in transferring from simulation to the real world. Recent work mainly focuses on domain randomization by generating diverse configurations in simulation \cite{Dogar-2012-7531,Tobin2017DomainRF,James2017TransferringEV}. Other approaches use pixel-level or feature-level domain adaptation \cite{DomainAdaptationBousmalis2018,Fang2018MultiTaskDA,Sim2RealJames2018}. At best these methods still require a significant amount of unlabelled real world data, which is still costly and time-consuming to collect and may not transfer to a new task.

\begin{figure}[t]
\centering
\hspace*{-0.1in}
\includegraphics[width=1.0\linewidth]{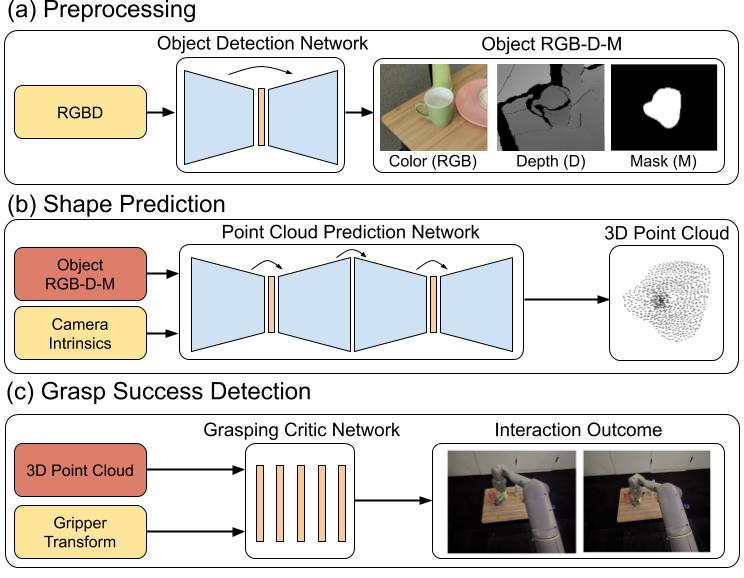} 
\caption{Architecture overview: (a) we use an object detection network to obtain object-centric color, depth and mask images; (b) Our point cloud prediction network allows us to generate a 3D point cloud of the detected object; (c) Finally, we use a grasping critic network to predict a grasp. 
}
\vspace{-2mm}
\label{fig:introduction}
\end{figure}

In contrast to end-to-end training frameworks, we consider learning visual structures as an intermediate representation for sim-to-real transfer.
More specifically, we embrace the recent advances in using deep neural networks to predict 3D structures (e.g., 3D voxel grids, point clouds, and triangle meshes) from single-view observation~\cite{FanSG17,Henderson2018LearningTG,varley2016shape}. Compared to 2D sensory input, 3D structure is known to be very useful for shape-based object manipulation such as grasping~\cite{goldfeder2009columbia,leon2010opengrasp,bohg2010learning}. For example, the geometric center of an object, is a shape feature useful for its localization and manipulation, which can be inferred directly from 3D structures~\cite{Xu2018PointFusionDS}.

Although shape prediction modules can be useful, it remains a non-trivial task to apply the existing work to robotic platform in the real world. First, inferring a depth image solely from a single-view RGB image based on traditional computer vision techniques introduces ambiguities. Recent learning-based work on image-to-depth prediction have demonstrated good performance on the depth prediction from a single RGB camera~\cite{Eigen2015,zhou2017unsupervised,Garg2016}, 
however, applications of these methods to robotics tasks are not yet well explored.
Second, even with additional depth channel as input, sim-to-real transfer is not straightforward. For example, depth sensors often inject arbitrary noise (e.g. when the object is dark or transparent, or when there is misalignment between depth and RGB) in the real world.

In this work, we design a novel shape prediction model that generates full 3D point clouds of an object from a single-view RGBD image sampled from a sequence containing multiple snapshots of the scene.
We further explore cross-view consistency as the self-supervision signal for training, as multiple input views share the same intermediate representation given camera transformations (e.g., rotating the 3D shape from one view to another).
Compared to other 3D representations such as voxels~\cite{wu2016learning,yan2016perspective} and triangle meshes~\cite{kato2018neural,wang2018pixel2mesh}, 3D point clouds are lightweight (i.e. low-dimensional) and free from aliasing artifacts under camera transformations. 

In summary, our contributions are: 
\begin{itemize}
\item we present a self-supervised shape prediction framework that reconstructs full 3D point clouds as representation for robotic applications;
\cutparagraphup
\item we show data-efficient and robust sim-to-real transfer using this self-supervised framework;
\cutparagraphup
\item we demonstrate the application of the predicted point cloud on the robotic task of table-top instance grasping using zero real-world grasping data.
\cutparagraphup
\end{itemize}

\cutsectionup
\section{Related Work}
\cutsectiondown

Learning to interact with objects is a wide and actively studied field of vision and robotics research. Many approaches are based on using visual features obtained from RGB or RGBD images to identify objects and grasping points~\cite{saxena2008robotic,montesano2012active,lenz2015deep,gualtieri2016high,kopicki2016one,osa2016experiments,pinto2016supersizing}. While early approaches focus on studying the problem of grasping based on traditional techniques, such as logistic regression or learning probabilities of grasp success~\cite{saxena2008robotic,montesano2012active}, more recent approaches often rely on deep neural networks to extract more nuanced features from images. This ranges from detecting objects~\cite{lenz2015deep} and their pose~\cite{gualtieri2016high}, to learning grasp types from kinesthetic demonstrations~\cite{kopicki2016one}, under gripper pose uncertainty~\cite{johns2016deep}, or following an unsupervised learning scheme~\cite{Grasp2Vec2018}. The effectiveness of deep neural networks is unparalleled, however, training the networks requires large-scale labeled datasets to generalize to unseen objects~\cite{pinto2016supersizing}. Other approaches for robotic grasping focus on identifying the grasp affordance of objects~\cite{dang2014semantic,katz2014perceiving} or by categorizing them according to their function~\cite{nikandrova2015category}.

Another line of research focuses on reconstructing objects and scenes as 3D triangle meshes, which in turn can be used to enable more informed robotic behavior~\cite{li2016dexterous,vahrenkamp2016part}. Reconstructing objects from incomplete scans is challenging. Various approaches exist to reconstruct the geometry of objects, while considering the many facets of the problem~\cite{6162880,7780981,ganapathi2018,dai2018scancomplete,Li:2015:DOR:2816723.2816762,SongYZCSF17}. 
Very recently, Henderson and Ferrari~\cite{Henderson2018LearningTG} introduced a network architecture to generate 3D meshes, while only providing single image supervision. At a higher level, our method is related to  recent work on deep 3D representation learning for robot-object interaction~\cite{eslami2018neural,yan2018learning,wang20183d}. Moreover, to facilitate the learning of 3D shape representations and the grasping of objects, recent efforts concentrate on curating large shape and grasp repositories~\cite{Chang2015ShapeNetAI,mahler2016dex,mahler2017dex}. 

It has been recognized that point sets can serve as an effective representation to obtain additional information of objects. 
Several recent approaches learn representations to generate point clouds of shapes~\cite{FanSG17,Achlioptas2018LearningRA,mrt18}.  
More closely targeted to robotics, Xu et al.~\cite{Xu2018PointFusionDS} and Wang et al.~\cite{Wang2019} propose fusion networks to extract pixel-wise dense feature embeddings for estimating 3D bounding boxes and 6D object pose respectively. These approaches focus on robotic vision, and unlike them we do not rely on the 6D object pose as a label for training. 

Reconstructing the geometry of an object allows identifying grasping points more precisely and thereby to control the grasp~\cite{goldfeder2009columbia,leon2010opengrasp,bohg2010learning}. To this end, Varley et al.~\cite{varley2016shape} and Yan et al.~\cite{yan2018learning} recently proposed to reconstruct shapes by learning geometry-aware representations based on 3D occupancy grids. Grids serve as an efficient representation but they obscure the natural invariance of 3D shapes under geometric transformations and only support to represent shapes at low resolution. Similar to our method these approaches also aim at reconstructing objects or object parts to enable more robust grasping. However, unlike them we focus on the self-supervised reconstruction of full 3D point clouds of objects. Point clouds enable a more robust sim-to-real transfer and thereby facilitate efficient training. 

Methods on sim-to-real transfer aim at training neural networks on simulated data with the goal to operate on real data at inference time. This is also known as domain adaptation, where a model is trained with data points of a source domain to then generalize to a target domain~\cite{7078994,Csurka2017DomainAF}. As simulated data can be generated efficiently with ground-truth labels, a number of methods focus on sim-to-real transfer by enhancing or synthesizing images~\cite{saxena2008robotic,Viereck2017LearningAV,Gualtieri2016HighPG}. More recently, Bousmalis et al.~\cite{DomainAdaptationBousmalis2018} introduced a generative adversarial network pipeline to enhance simulated data to significantly reduce the number of real-world data samples for a grasping task. Fang et al.~\cite{Fang2018MultiTaskDA} propose a multi-task domain adaptation framework composed of three grasp prediction towers for instance grasping in cluttered scenes. Finally, James et al.~\cite{Sim2RealJames2018} introduce a two-stage generator pipeline to translate simulated and real images into a canonical representation to realize sim-to-real transfer. While existing work mostly focuses on introducing architectures to operate on images, we argue that point clouds of objects serve as an effective representation to facilitate sim-to-real transfer. 


\begin{figure*}[t]
\centering
\hspace*{-0.1in}
\includegraphics[width=\linewidth]{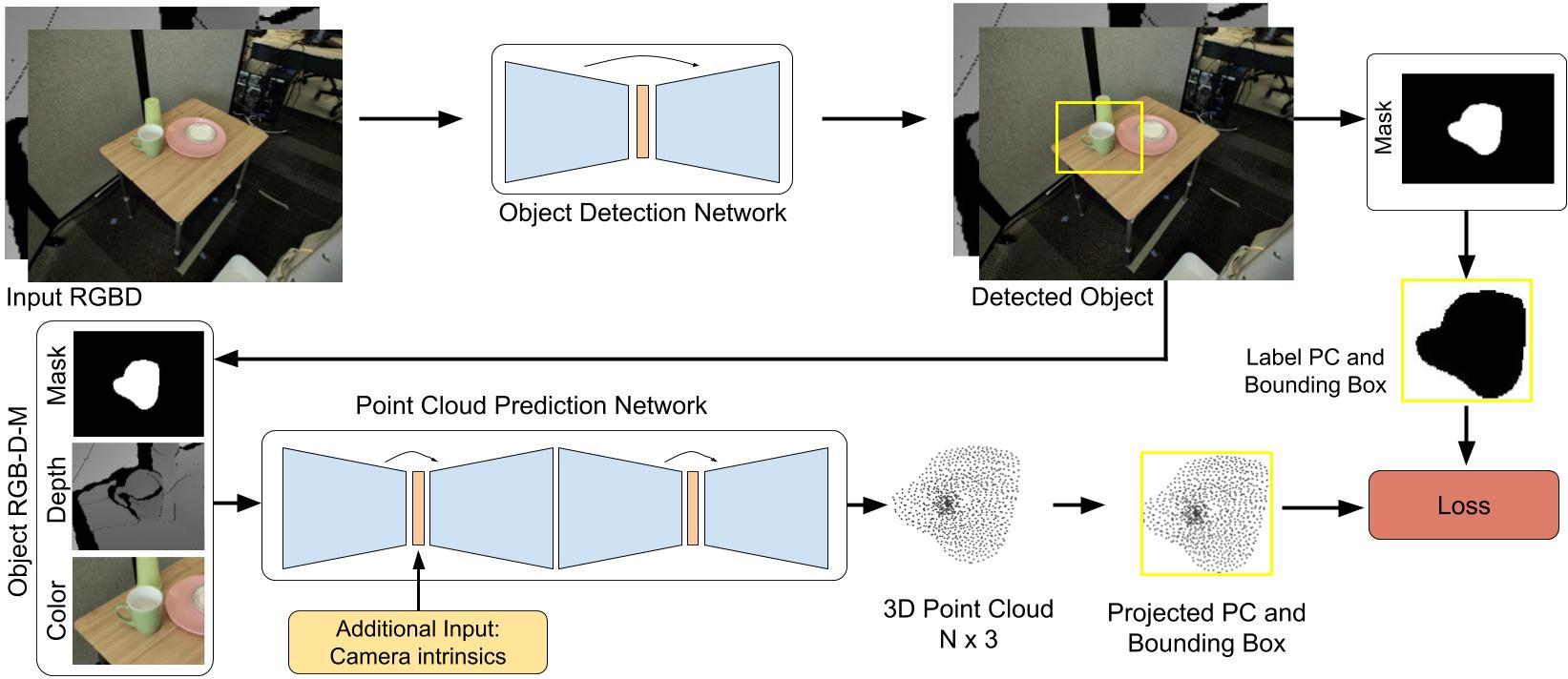} 
\vspace{2mm}
\caption{Overview of our object detection and point cloud prediction networks: we detect an object and obtain its cropped color and depth images along with a binary mask based on an object detection network. We then use the 5 channels (RGBD-M) to train a point cloud prediction network that allows us to predict a 3D point cloud (PC) of the detected object. Using the depth image of the object and its mask, we compute a ground truth label for object depth. The network is trained against an image-based loss of the projected 3D point cloud and the ground truth mask of the object. 
}
\vspace{-3mm}
\label{fig:network_architecture}
\end{figure*}

\cutsectionup
\section{Methods}
\cutsectiondown

Given a set of RGBD observations, we aim at learning an intermediate representation that is \textbf{compact}, \textbf{domain-invariant}, semantically \textbf{interpretable}, and directly \textbf{applicable} for object manipulation.
In particular, we use a point cloud of a target object of interest as our intermediate representation for learning due to the following reasons: 
(1)~Point clouds are more lightweight and flexible compared to other 3D representations such as voxel grids and triangle meshes.
(2)~Point clouds describe the full 3D shape of a target object, which is inherently invariant to surface textures or environmental conditions.
(3)~A point cloud representation can directly be used to localize objects in the scene, hence simplifying the tasks when training a policy.
Our approach includes two steps: 1) Learning a domain-invariant representation using visual observations from simulation and real world, which will be described in Section~\ref{sec:representation_learning}, and 2) Learning an object manipulation policy such as grasping using the representation from step 1, which will be described in Section~\ref{sec:grasping_policy}.

\subsection{Learning domain-invariant representation}
\label{sec:representation_learning}

Given a set of RGBD observations $\left\{\mathcal{O}_1, \mathcal{O}_2, \cdots, \mathcal{O}_N\right\}$, where $\mathcal{O}_n \in R ^ {h \times w \times 4}$ is an individual observation which could come from simulation or real world, our  goal is to learn a domain-invariant point cloud representation $\mathcal{P}$ reflecting the 3D geometry of the target object. These $N$ observations can be easily obtained by using a mobile manipulator moving around and taking snapshots of the workspace from different angles. We assume each object may be present in more than one snapshot as it allows to reconstruct the 3D geometry of the object better. However, we do not impose any explicit constraint on the number of snapshots that the object should be present in. 
Please note that the depth values from RGBD observations forms a 2.5D representation of the objects (e.g., visible part subject to noise) and thus do not provide the full 3D geometry. Furthermore, there is a reality gap between the depth values in simulation and real world, hence making a policy that is solely trained in simulation quite ineffective in the real world.

\paragraph{Self-supervised labeling.}
While target point clouds for supervised learning of a deep network can be easily obtained in simulation, this task becomes notoriously costly and time-consuming for real data. Furthermore, the presence of noise and unmodeled nonlinear characteristics in a depth sensor make the learning harder, especially in the context of transfer learning. To address this challenge, we base our framework on the recent work of learning to reconstruct 3D object geometry using view-based supervision with differential re-projection operators \cite{jiang2018gal,Liu2019SoftRD}.

We represent a point cloud $\mathcal{P}$ of an object $\mathcal{O}$ as a set of $K$ points $\mathcal{P} = \{\mathbf{p}_k = (x_k, y_k, z_k) | 1 \le k \le K\}$, where $x_k$, $y_k$, $z_k$ are coordinates regarding $k$-th point $\mathbf{p}_k$ along xyz axis, respectively. 
Without loss of generality, we assume the point cloud coordinates are defined in the camera frame.
We assume the ground-truth point cloud annotation is not directly available in the real world data and thus use multi-view projections as the supervision signal.
More specifically, we use the camera intrinsic matrix $E$ to obtain the 2D projection $\mathcal{M}$ in the image space from the point cloud $\mathcal{P}$ (e.g., homogeneous coordinate $(u_k, v_k, 1)$ is projected from $(x_k, y_k, z_k)$):
\begin{subequations}  
\begin{align}
(u_k, v_k, 1)^\top &\sim E \; (x_k, y_k, z_k)^\top \\
\mathcal{M} &= \left\{(u_k, v_k)\right\}_{k=1}^K
\end{align}
\label{eqn:projection}
\end{subequations}  

For localization, the corresponding tight bounding box can be derived from the 2D projection: $\mathcal{B} = (u^\text{mid}, v^\text{mid}, w, h)$, where $u^\text{mid}$, $v^\text{mid}$, $w$, $h$ represents the bounding box center and size, respectively.

We collect $N$ RGBD snapshots from various scenes in simulation and real world by moving a mobile manipulator around the workspace. For the real world dataset we use Mask-RCNN \cite{he2017maskrcnn} to detect object bounding boxs $\mathcal{B}^n$ and their associated mask at each frame. For the simulation dataset bounding boxes can directly be obtained. Note that it is quite common that multiple objects may be present in each snapshot. We denote the data associated to the $m$-th object in the $n$-th frame by $(.)^{m, n}$. We also denote the number of objects in the $n$-th frame by $C_n$. Next, we use the mask for each object to extract its associated depth values from the depth channel in each observation and then use the camera intrinsic matrix $E^n$ to obtain $\mathcal{M}^{m, n}$ from the depth values. At the end of this step, we obtain $\mathcal{M}^{m, n}$ and $\mathcal{B}^{m, n}$ for all $1 \leq n \leq N$ and $1 \leq m \leq C_n$.

Our deep net provides an estimate of the point cloud $\hat{\mathcal{P}}$ as the output which can be used to determine $\hat{\mathcal{M}}^{m, n}$ and $\hat{\mathcal{B}}^{m, n}$ using Eq. \ref{eqn:projection}. Then we define the loss function for training the domain invariant point cloud representation as follows:
\begin{align}
\mathcal{L}_\theta
	= \sum_{n=1}^N \sum_{m=1}^{C_n} \Big( \lambda^\mathcal{B} \mathcal{L}_\theta^\mathcal{B}(\hat{\mathcal{B}}^{m,n}, \mathcal{B}^{m, n}) & + \nonumber \\
	& \hspace{-3cm} + \lambda^\mathcal{M} \mathcal{L}_\theta^\mathcal{M}(\hat{\mathcal{M}}^{m, n}, \mathcal{M}^{m,n}) \Big) + \lambda^\theta \|\theta\|
\label{eqn:formulation_multi}
\end{align}
\noindent where $\lambda^{\mathcal{B}}$, $\lambda^{\mathcal{M}}$, and $\lambda^{\theta}$ are weighting coefficients, $\mathcal{L}_\theta^\mathcal{B}$ is the Huber loss between the estimated and labeled bounding box, and $\mathcal{L}_\theta^\mathcal{M}$ is the projected point-cloud prediction loss based on \cite{jiang2018gal}. We extend their method in the following way: we do not rely on obtaining a full 3D point cloud as this is more difficult to accomplish in real-world environments. Moreover, while \cite{jiang2018gal} uses synthetic 3D point clouds for training to reconstruct normalized shape centered at the origin, our goal is to reconstruct shapes at real-world scale. 

\textbf{Network architecture.}
Figure~\ref{fig:network_architecture} illustrates the network architecture we use for shape prediction. Similarly to \cite{jiang2018gal}, we use a network that is composed of several encoder-decoder modules \cite{FanSG17} and a fully-connected layer to predict the point clouds. However our proposed architecture is different from \cite{jiang2018gal} in three ways in order to make it applicable for real world robotics setting: (1) we use object masks as an additional input channel to handle situations when multiple objects are present in the scene (a very common situation in robot settings). Therefore, the number of input channels to our network is five (RGBD-Mask), (2) we introduce a dynamic image cropping step on the RGBD-M channels. This allows to get a more focused view of the target instance, and (3) to account for the dynamic cropping, we add an additional input right after the encoder to provide the network with the adapted camera intrinsic characteristic resulted from cropping. 

\subsection{Learning Point cloud-based Grasping Policy}
\label{sec:grasping_policy}

In this section we describe how the learned point cloud representation can be used to perform table-top instance object grasping. In this work, we use a critic network $\mathcal{G}(\mathcal{P}, \mathbf{s}, \mathcal{T}_R^C)$ to predict the probability of the success for a sample table-top grasp $\mathbf{s} \in R^4$ based on the predicted point cloud $\mathcal{P}$ for the target object, and the transformation from the robot base to the camera frame. In our setup the sample $\mathbf{s} = (\mathbf{p}, \psi)$ is composed of the 3D gripper position $\mathbf{p}$ with respect to the robot base and the gripper yaw rotation~$\psi$. 

\begin{figure}[t]
\centering
\includegraphics[width=\linewidth]{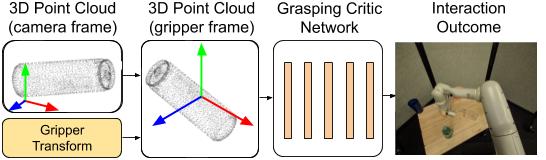} 
\caption{
Overview of our point cloud-based grasping network.
The point clouds are first transformed to the gripper frame, followed by a classification network adapted from PointNet~\cite{qi2017pointnet} architecture with 5 fully-connected layers in total.
%
}
\vspace{-4mm}
\label{fig:grasping_critic}
\end{figure}

\paragraph{Network architecture.}
Figure~\ref{fig:grasping_critic} shows the architecture of our grasp prediction network. For the preprocessing, we first transform the point cloud $\mathcal{P}$ to the proposed grasp frame using:
\begin{equation}
    \mathcal{P}_s = \mathcal{T}_s^R \mathcal{T}_R^C \mathcal{P}
\end{equation}
\noindent where $\mathcal{T}_s^R$ can be directly calculated based on the sample grasp pose $\mathbf{s}$. We then shuffle the order of points in $\mathcal{T}_s^R$ to allow the network to adapt to variations in the order of point clouds. 

Our grasping critic network is derived from PointNet~\cite{qi2017pointnet} architecture, which is composed of 4 fully-connected layers each followed by a \texttt{ReLU} activation function with \texttt{BatchNorm}. The last layer is linear and reduces the output size to 1. It is followed by a \texttt{sigmoid} activation function to provide the grasp success.

\paragraph{Generating grasping data for training.}
We collected data for training the grasp policy only in simulation. We use a heuristic grasping policy for the data collection as follows: (1) Compute the center of volume of the object $\mathbf{p}_\text{obj}$ based on the predicted point cloud, (2) Set the translation part of the grasp pose to $\mathbf{p}_\text{obj}$ plus some random noise $\mathbf{p}_\epsilon \in R^3$, i.e. $\mathbf{p}_\text{grasp} = \mathbf{p}_\text{obj} + \mathbf{p}_\epsilon$, (3) Randomly draw a yaw angle from a uniform distribution in the range $[-\pi/2, \pi/2]$.

We then evaluate the grasp success by moving the arm first to a pre-grasp pose $\mathbf{s}^*$, where $\mathbf{s}^*$ is a pose exactly above $s$ with some height difference constant, i.e. $\mathbf{s}^*-\mathbf{s} = (0, 0, \Delta{h}, 0)$. This allows to properly pose the robot end-effector with respect to the object before attempting the grasp. Then the robot is moved to pose $\mathbf{s}$, and finally we command the robot to close its parallel-jaw gripper. The robot is then commanded to lift the object by moving back to $\mathbf{s}^*$. Then the grasp success is evaluated by checking whether the object is moved above the table. This evaluation can be done easily since we have access to the ground truth object pose in simulation. The training data is collected by running simulation robots in parallel and stored for training the off-policy grasping network.

\paragraph{Runtime evaluation.}
One of the main advantages of using point clouds is that we directly have access to the 3D location of an object in the scene. Given this knowledge, the simplest approach to infer a grasp pose is to randomly sample several grasp pose candidates around the object pose (similar to our data collection procedure), and then take the command with the highest grasp success probability. However, to get better results, we use the cross-entropy method (CEM)~\cite{rubinstein_cem} to find the most suited grasp pose by using a simple derivative-free optimization technique. Using CEM, at each iteration we sample a batch of $N_\text{CEM}$ grasp candidates. Then evaluate these samples and pick $N_\text{elite}$ best ones ($N_\text{elite} \ll N_\text{CEM}$). Next, we fit a Gaussian distribution to $N_\text{elite}$ and then sample a new batch of size $N_\text{CEM}$. We repeat this process a few times until either a grasp candidate above the desired success probability threshold $\alpha$ is achieved or when the CEM reaches the maximum number of allowed iterations. In our implementation we use $N_\text{CEM}=100$, $N_\text{elite}=10$, $\alpha=0.9$ and max number of iterations 3.

Note that since we are directly sampling the grasp pose, we can directly impose additional constraints to our sampling strategy (e.g., limiting samples to the desired workspace as well as removing kinematically infeasible samples).

\cutsectionup
\section{Experiments}
\cutsectiondown

In the following we provide details on our real and simulated datasets and discuss results on shape prediction and grasping performance with a real robot. 

\subsection{Datasets.}

\begin{table}[t!]
\centering
\caption{Statistics of shape prediction datasets.}
\scalebox{0.8}{
\begin{tabular}{l||c|c|c}
\hline
Dataset (split) & \# objects & \# episodes & \# object-centric seqs \\
\hline\hline
ShapeNet & 1,345 & 38,653 & 189,424 \\
\hline\hline
Kitchenware (train) & 69 & 37,568 & 142,293 \\
Kitchenware (test) & 20 & 3,092 & 11,956 \\
\hline\hline
Real (train) & 110 & 534 & 38,301 \\
Real (val) & 34 & 49 & 2,720 \\
Real (test) & 43 & 291 & 18,462 \\
\hline	
\end{tabular}
}
\label{tab:table-shape-dataset}
\end{table}

\begin{table}[t!]
\centering
\caption{Statistics of simulated grasping datasets.}
\begin{tabular}{l||c|c}
\hline
Dataset (split) & \# objects & \# episodes \\
\hline\hline
Kitchenware (train) & 69 & 358,286\\
Kitchenware (test) & 20 & 2,111\\
\hline\hline
Procedural & 1,000 & 633,069\\
\end{tabular}
\label{tab:table-grasp-dataset}
\vspace{-3mm}
\end{table}

\paragraph{Shape prediction datasets.}
We use both real-world data and data generated from simulation for learning domain-invariant 3D shapes.
As shown in Table~\ref{tab:table-shape-dataset}, our real-world dataset contains 187 distinct object instances in total, covering 8 object categories: \textit{balls}, \textit{bottles} \textit{cans}, \textit{bowls}, \textit{cups}, \textit{mugs},  \textit{wine glasses}, and \textit{plates}.
To obtain sufficient objects for training our shape prediction model, we use additional CAD models from ShapeNet~\cite{Chang2015ShapeNetAI} and Kitchenware~\cite{yan2018learning}.
We use an off-the-shelf simulator PyBullet~\cite{coumans2019} to create virtual scenes in the simulation.
For ShapeNet, we select $1,345$ objects from 6 categories: \textit{bags}, \textit{bottles}, \textit{bowls}, \textit{cans}, \textit{jars}, and \textit{mugs}, which are similar to the objects of our real-world dataset.
Moreover, we use Procedural dataset~\cite{DomainAdaptationBousmalis2018} containing $1,000$ procedurally generated object shapes for evaluation.

\paragraph{Simulated grasping datasets.}
Assuming our shape representation is domain-invariant, we only generate grasping data in simulation.
As shown in Table~\ref{tab:table-grasp-dataset}, we use the Kitchenware and Procedual dataset~\cite{DomainAdaptationBousmalis2018} for training.
Additionally, we use a subset of the Kitchenware dataset for grasping evaluation in simulation.
These are held-out objects which have never been used for learning the domain-invariant 3D shapes.

\begin{figure}[t]
\centering
\includegraphics[width=\linewidth]{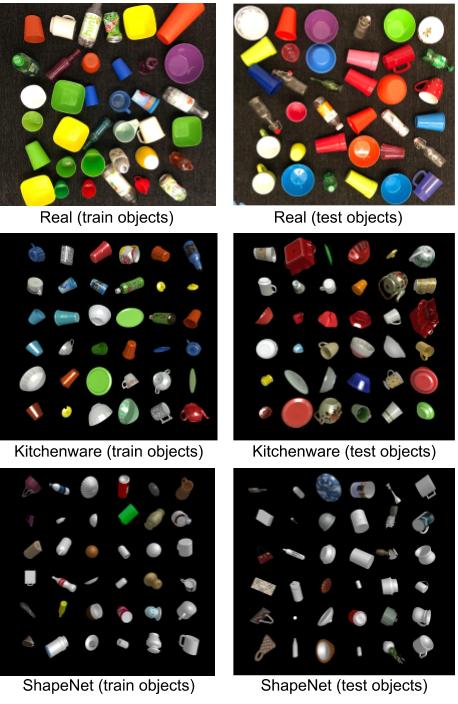} 
\caption{
Overview of the dataset used for learning the domain-invariant shape prediction model. We visualize the object instances used for training the point cloud prediction model in the real world and simulation (e.g., Kitchenware dataset and ShapeNet subset) from top to bottom.
}
\label{fig:data_objects}
\vspace{-2mm}
\end{figure}

\begin{figure}[t]
\centering
\includegraphics[width=\linewidth]{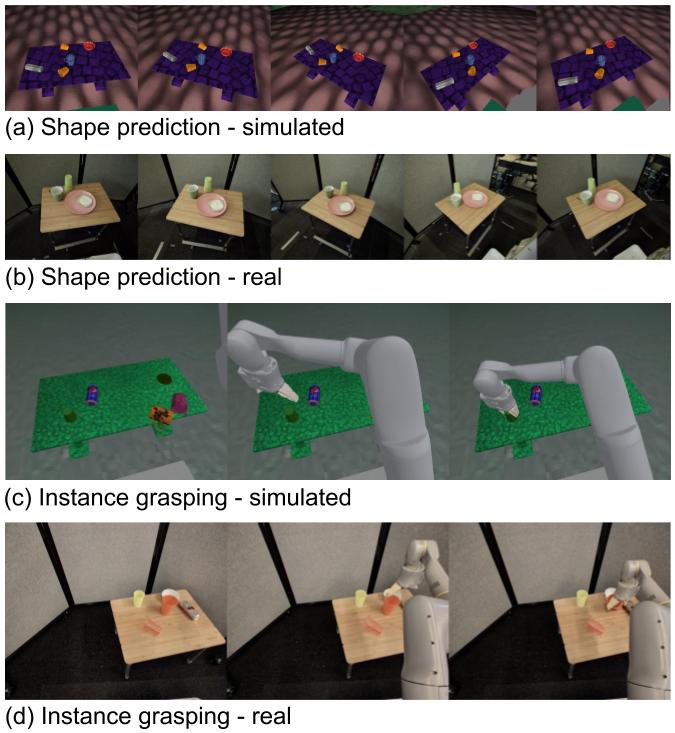} 
\vspace{-4mm}
\caption{Examples of data collection for shape prediction in the simulated (a) and real environment (b) and for instance grasping in the simulated (c) and real (d) setup. 
}
\vspace{-2mm}
\label{fig:data_collection}
\end{figure}

\paragraph{Scene construction and pre-processing.}
As illustrated in Figure~\ref{fig:data_collection}, in both real and simulation environment, we put a table in front of the mobile manipulator and randomly set the table height between 30 and 50 centimeters.
In the simulation, we perform additional randomization over the table location and object textures (e.g., texture pattern of the background, color of the table).
We place 4 to 5 objects on the table with arbitrary location and orientation.
Given the scene arrangement, we place the mobile manipulator at 5 different angles looking at the table from one side.
For each viewpoint, the robot takes one snapshot containing RGB and depth images with a resolution of $512 \times 640$.

For object detection and segmentation we used Mask R-CNN~\cite{he2017maskrcnn}. Given an image of a scene, Mask-RCNN detects and segments objects above a threshold, which generates bounding boxes and segmentation masks for each instance of an object.
For our setup we are interested in detecting four types of object categories: \textit{bottles}, \textit{wine glasses}, \textit{cups}, and \textit{bowls}. 
If bounding boxes overlap we use an IOU threshold of 0.5 to remove duplicate objects as part of the non-maximum suppression step. 
For each detected object, we generated a cropped image and re-scale it to size $192 \times 256$.
For multiple snapshots taken at the same scene, we associate objects across snapshots based on the additional depth information.
For the simulated environment, we introduce $4$ virtual viewpoints to obtain a full 360 degree capture of the scene.
The robot looks at the table from different viewpoints. We then extract object point clouds based on the object detection inference results.

\begin{figure}[t]
\centering
\hspace*{-0.1in}
\includegraphics[width=1.03\linewidth]{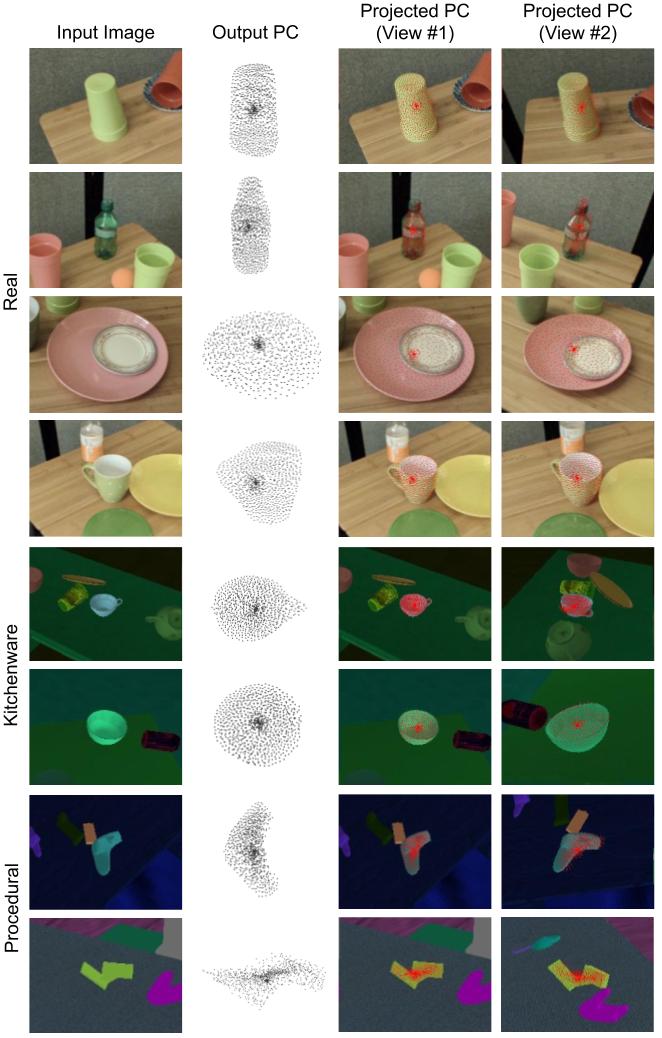} 
\caption{Visualizations of point clouds generated with our point prediction network. From left to right: the input image, the generated 3D point cloud, and two different views with the projected point cloud as an overlay on the detected object. Our method allows us to produce meaningful point clouds from real and simulated test data (Kitchenware, Procedural). 
}
\label{fig:shape_vis}
\end{figure}

\subsection{Experimental Results}
\paragraph{Evaluation metrics.} We consider both shape prediction performance and grasping performance for evaluation.
For shape prediction, we use bounding boxes from the simulation or Mask-RCNN as ground-truth.
We project the predicted 3D point cloud to 2D and evaluate the overlap of the prediction and the ground truth.
To measure the performance, we run the instance grasping and compute the grasping success rate.
Please note that our method relies on object detection based on Mask-RCNN, which may fail to reliably detected objects. While this can consequently also result in grasping failure, we consider this as an orthogonal problem that is outside the scope of this work.  

\paragraph{Evaluation on shape prediction.}

To decide whether our shape prediction model is able to generate meaningful point clouds from a single RGBD observation, we conducted  qualitative and quantitative evaluations on predicted 3D point clouds.
In Figure~\ref{fig:shape_vis} we illustrate that our shape prediction model is able to successfully predict 3D point clouds. The point clouds (second column) are shown from the view of the input image.
One advantage of point clouds as representation is that they can be easily transformed from one view to another.
The right-most column in Figure~\ref{fig:shape_vis} shows the projected 2D points still align with the object when looked at from another view.
This also addresses the concern that our model does not learn the trivial mapping from input object mask.
It is worth noting that our shape prediction model also generalizes to unseen categories from \textit{Procedural} dataset (see the last two rows of Figure~\ref{fig:shape_vis}).

For a quantitative evaluation, we use more than 10K image patch sequences containing the object instances of the test set of both \textit{Kitchenware} and our real data.
We compute the averaged IOU of the 2D projections with ground-truth masks (from the simulation or Mask-RCNN) and summarize the results in Table~\ref{tab:table-shape-iou}.
We also conducted an ablation study on the number of viewpoints used during training.
The results are reported in Table~\ref{tab:table-shape-iou}. The model performs poorly when only one view is used for supervision, however, the  performance increases when multiple views are provided.

\begin{table}[t!]
\centering
\caption{
Shape Prediction IOU on unseen objects.
}
\scalebox{1.0}{
\begin{tabular}{l||c|c|c|c}
\hline
Dataset $/$ \# views & 1 & 2 & 4 & full \\
\hline\hline
Kitchenware & 0.188 & 0.659 & 0.803 & 0.803 \\
\hline
Real & 0.186 & 0.492 & 0.625 & 0.626\\
\hline	
\end{tabular}
}

\label{tab:table-shape-iou}
\end{table}

\begin{table}[t!]
\centering
\caption{
Grasping success rate on unseen objects.
}
\scalebox{1.0}{
\begin{tabular}{l||c|c}
\hline
Dataset & 2.5D shape & Reconstructed 3D shape\\
\hline\hline
Kitchenware & 68\% & 64\% \\
\hline
Real & 51\% & 61\% \\
\hline	
\end{tabular}
}
\label{tab:table-grasp-perf}
\vspace{-2mm}
\end{table}

\begin{figure*}[t]
\centering
\includegraphics[width=1.0\linewidth]{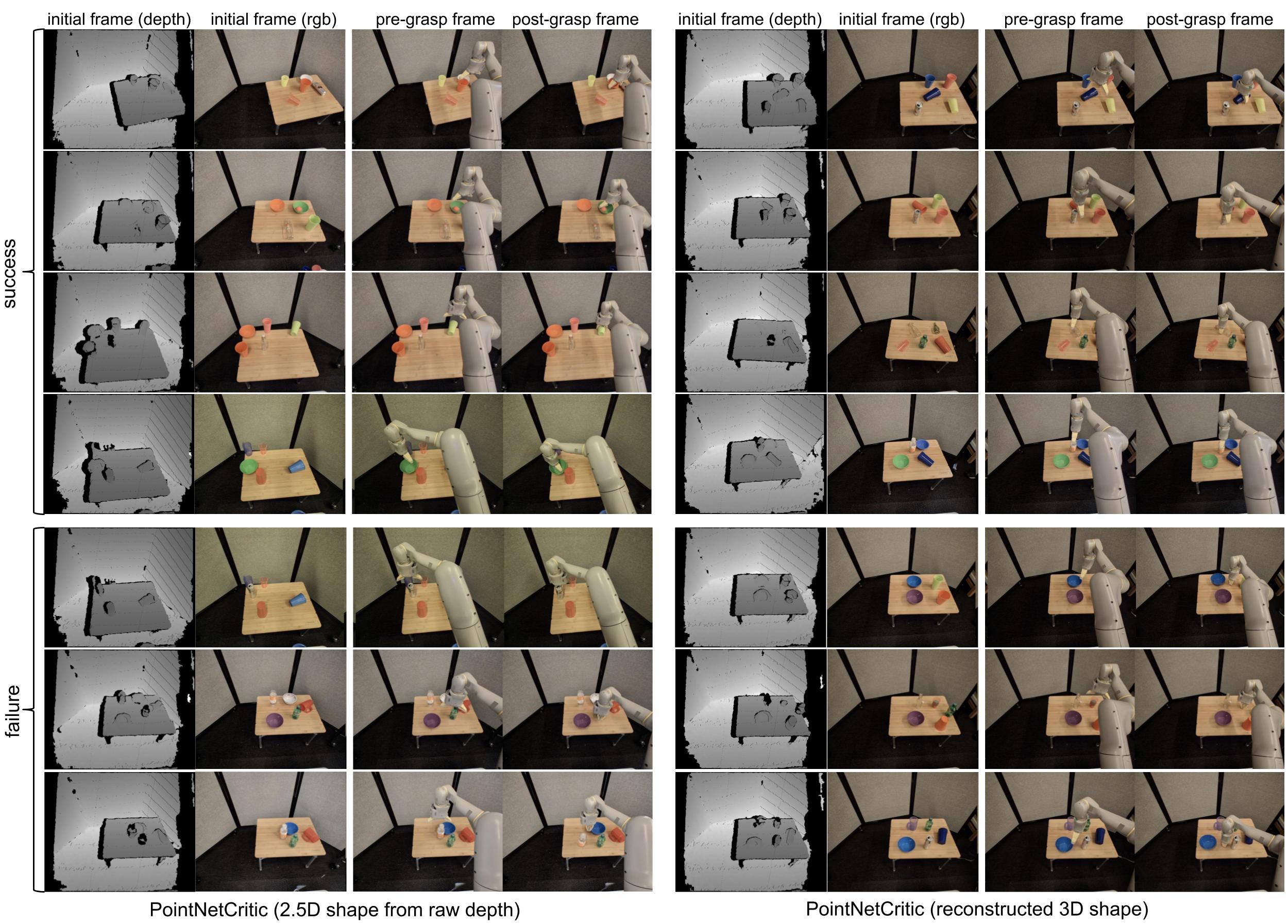} 
\caption{Grasping sequence evaluation: we visualize the real world grasping sequences for the baseline model (left) and our model (right).
For both, the first two columns show the depth and RGB image before we run grasping trials.
The third column illustrates the pre-grasp state where the robot arm has been moved on top of the target instance.
The last column shows the final state after the grasping has been executed.
}
\label{fig:grasp_eval_vis}
\end{figure*}

\paragraph{Evaluation on instance grasping.}
Finally, we evaluated whether the learned critic model is able to guide the CEM policy for grasping on both simulated and real data (the dataset contains $20$ and $43$ unseen objects respectively).
Specifically, we execute 100 grasping trials in the real world and in the simulation. We report the average grasping success rate in Table~\ref{tab:table-grasp-perf}.
During execution, we follow the same protocol used for generating the simulation dataset for training (see Figure~\ref{fig:data_collection}).
The robot is expected to pick up one specific object from the table and drops it into a bin.
If the grasp was not successful, we manually remove the object from the table.
This prevents the model from repeatedly grasping the same object.

As reported in Table~\ref{tab:table-grasp-perf}, CEM policy guided by our critic model achieves 64\% instance grasping success rate in simulation and 61\% in the real world execution with zero real data for training.
This is a significant improvement over previous work that aims at sim-to-real transfer at image-level (e.g., \cite{DomainAdaptationBousmalis2018} achieved 23.53\% indiscriminative grasping success trained only with simulated data).

To evaluate the performance of grasping success with our reconstructed 3D point clouds as representation for sim-to-real transfer, we learn a baseline critic model that only uses raw 2.5D sensor depth as the input.
As reported in Table~\ref{tab:table-grasp-perf}, the CEM policy guided by the baseline critic model achieves slightly higher grasping success rate in simulation (e.g., 68\% vs. 64\%), as it is trained on the clean ground-truth point cloud.
However, policy guided by the baseline model suffers severely from domain shift, as the success rate drops from 68\% to 51\% when applied in the real-world environment.
As we used the same architecture and training set for both our model and the baseline, we believe the performance gap is the result of domain shift (e.g., the unmodeled noise from the depth input, the noise introduced by our Mask-RCNN model).
Due to our domain-invariant 3D representation, the policy based on our critic model achieved 10\% higher success rate compared to the policy based on raw depth from the sensor.
These results are shown in Table~\ref{tab:table-grasp-perf} and Figure~\ref{fig:grasp_eval_vis}.
To summarize, the improvement illustrates clear advantages of our compact and geometry-aware domain-invariant representation.

\section{Conclusions and Future Work}

In this work we presented a novel self-supervised approach that learns to perform table-top instance grasping of objects using no real world grasping data.
The proposed framework consists of a shape prediction model that learns a domain-invariant 3D point cloud representation of objects by operating on single RGBD images containing the same object from different viewpoints.
Moreover, the 3D point cloud is further utilized to perform instance grasps based on a critic model learned with simulated grasping data only.

Experimental results have demonstrated that (1) our shape prediction model is able to learn domain-invariant 3D shapes in real world settings from single RGBD image observations with self-supervision; (2) the policy guided by our representation generalizes significantly better in the real world compared to previous state-of-the-art based on end-to-end training and a policy based on a 2.5D shape representation.

In a broader context, our shape-aware representations provide a better understanding of objects and thereby have the potential to enable more robust robotic behavior towards grasping and other manipulation tasks.
As avenues for future work, it would be interesting to explore the potential of such representations with respect to more diverse object categories or end-effector configurations as well as a larger number of tasks.

\newpage
\bibliographystyle{ieee}
\bibliography{references}

\end{document}